\documentclass[10pt,twocolumn,letterpaper]{article}

 \usepackage[pagenumbers]{cvpr} % To force page numbers, e.g. for an arXiv version

\definecolor{cvprblue}{rgb}{0.21,0.49,0.74}
\usepackage[pagebackref,breaklinks,colorlinks,allcolors=cvprblue]{hyperref}

\def\paperID{*****} % *** Enter the Paper ID here
\def\confName{CVPR}
\def\confYear{2026}

\title{TSRE: Channel-Aware Typical Set Refinement for Out-of-Distribution Detection}

\author{Weijun Gao \qquad Rundong He \qquad Jinyang Dong \qquad Yongshun Gong\thanks{Corresponding author}\\
Shandong University\\
}
\begin{document}
\maketitle
\begin{abstract}
Out-of-Distribution (OOD) detection is a critical capability for ensuring the safe deployment of machine learning models in open-world environments, where unexpected or anomalous inputs can compromise model reliability and performance. Activation-based methods play a fundamental role in OOD detection by mitigating anomalous activations and enhancing the separation between in-distribution (ID) and OOD data. However, existing methods apply activation rectification while often overlooking channel's intrinsic characteristics and distributional skewness, which results in inaccurate typical set estimation. This discrepancy can lead to the improper inclusion of anomalous activations across channels. To address this limitation, we propose a typical set refinement method based on discriminability and activity, which rectifies activations into a channel-aware typical set. Furthermore, we introduce a skewness-based refinement to mitigate distributional bias in typical set estimation. Finally, we leverage the rectified activations to compute the energy score for OOD detection. Experiments on the ImageNet-1K and CIFAR-100 benchmarks demonstrate that our method achieves state-of-the-art performance and generalizes effectively across backbones and score functions.
\end{abstract}    
\section{Introduction}
Deep neural networks have been widely applied across various domains. However, when there is a distribution shift between the training and test sets, models may fail to handle inputs from the test set, potentially leading to system failures \cite{sun2022out, guo2017calibration, lakshminarayanan2017simple}. For example, an autonomous vehicle may be unable to recognize objects on roads that were not present in its object detection model, which could result in accidents \cite{han2022towards, michaelis2019benchmarking, joseph2021towards}. This problem is known as Out-of-Distribution (OOD) detection. To address this issue and ensure the safety of AI applications, researchers have been actively investigating OOD detection in recent years.

Existing OOD detection methods can be categorized into two main types:  density-based methods \cite{DBLP:conf/icml/ZhouL21, DBLP:conf/iclr/JiangSY22,zhang2023understanding} and classification-based methods \cite{DBLP:conf/iclr/HendrycksG17,mintun2022broad, che2023nado}. Since density-based methods generally underperform compared to classification-based methods \cite{DBLP:journals/corr/abs-2209-08590}, this paper focuses on the latter. Classification-based methods can be further divided into training-time \cite{DBLP:conf/iclr/DuWCL22,  DBLP:conf/cvpr/HeHLY22,zhu2024croft,xu2024out,sharifi2024gradient} and test-time methods, with the latter being more convenient as they do not require retraining. Test-time OOD detection methods can be classified into six types: confidence-based methods \cite{DBLP:conf/iclr/HendrycksG17, wang2023vim}, feature-based methods \cite{sastry2020detecting}, distance-based methods \cite{lee2018simple,liu2023detecting,liu2023fast, tao2023non}, gradient-based methods \cite{huang2021importance, li2024rethinking}, pruning-based methods \cite{sun2022dice, fort2021exploring} and rectified-activation-based methods. Rectified-activation-based methods have become a widely adopted class of test-time OOD detection methods, effectively mitigating anomalous activations by rectifying them into typical sets. This rectification improves ID-OOD separability by reducing overconfidence on OOD samples and underconfidence on ID samples \cite{zhu2022boosting}. Our work focuses on the last category: rectified-activation-based methods.

Most activation-based methods, such as BATS \cite{zhu2022boosting}, overlook channel-specific differences in Out-of-Distribution (OOD) detection. Although LAPS \cite{he2024exploring} takes into account the differences between channels, it only rectifies activations based on global-local statistical disparities, without considering the characteristics of the channel. Furthermore, activation-based methods make the assumption that activations across channels follow a Gaussian distribution. Our experimental results demonstrate that this assumption is invalid, and such distributional skewness can lead to inaccurate typical set estimation.

To address this limitation, we propose a channel-aware 
\underline{T}ypical 
\underline{S}et 
\underline{RE}finement (\textbf{TSRE})
 method that systematically rectifies feature activations. Unlike previous methods that apply uniform rectifications and neglect the distinct characteristics of individual channels, TSRE adaptively refines typical set boundaries by explicitly incorporating the discriminability and activity of each channel. To further mitigate bias introduced by incorrect distributional assumptions, TSRE refines the boundaries of the typical set based on skewness and rectifies anomalous activations into their corresponding typical sets. These rectified activations are then used to compute an energy score for OOD detection. Experiments on the ImageNet-1K and CIFAR-100 benchmarks demonstrate that our method achieves state-of-the-art performance and generalizes effectively across backbones and OOD scoring functions.

In summary, our contributions are as follows:
 
\begin{itemize}

\item We propose a channel-aware typical set refinement method that leverages each channel's discriminability and activity, addressing the underutilization of channel characteristics in prior work.
\item To further improve the accuracy of typical set estimation, we introduce a skewness-based refinement method to correct the distribution and mitigate skewness-induced boundary shifts, improving refinement accuracy and reducing distributional skewness.
\item We conduct comprehensive experimental evaluations on ImageNet-1K and CIFAR-100 benchmarks. The results demonstrate that our method outperforms existing methods and can be generalized to other architectures and OOD scores.

\end{itemize}

\section{Related Work}
Test-time Out-of-Distribution (OOD) detection has become a widely adopted solution due to its efficiency and practicality, as it does not require retraining the model. It is particularly advantageous in privacy-sensitive scenarios where fine-tuning on proprietary data is not feasible. These methods can be broadly categorized into six groups: confidence-based, feature-based, distance-based, gradient-based, pruning-based, and rectified-activation-based methods. \textbf{Confidence-based methods} estimate the likelihood of a sample being in-distribution by evaluating the prediction confidence of a pre-trained model. For example, Maximum Softmax Probability (MSP) \cite{DBLP:conf/iclr/HendrycksG17} considers the maximum softmax output as an OOD score, while ODIN \cite{DBLP:conf/iclr/LiangLS18} enhances this method by applying input perturbations and temperature scaling to the softmax logits. The Energy-based method \cite{liu2020energy} further enhances separability by using the energy score instead of softmax scores. \textbf{Feature-based methods} detect OOD samples based on internal representations. GRAM \cite{sastry2020detecting} computes the Gram matrix of intermediate features to capture spatial correlations, while SHE \cite{zhang2023out-of-distribution} leverages the associative memory mechanism of Hopfield networks. \textbf{Distance-based methods} are built on the assumption that OOD inputs lie farther from the training distribution in feature space. The Mahalanobis \cite{lee2018simple} computes distances between test samples and the class centroids of the training set. \textbf{Gradient-based methods} \cite{huang2021importance} use properties of the input gradients, such as gradient norms, to discriminate between ID and OOD data, based on the hypothesis that OOD inputs induce different gradient behaviors. \textbf{Pruning-based methods} prune the weights of the model to address overconfident predictions on OOD data. DICE \cite{sun2022dice}, for instance, prunes the output layer to reduce sensitivity to spurious activations. While effective in many cases, the above methods often fail to address abnormal activations that may cause unreliable predictions. Such anomalies can lead to inflated confidence on OOD inputs or subdued confidence on ID data, blurring the decision boundary. \textbf{Rectified-activation-based methods} seek to improve the separability between ID and OOD data by rectifying activations. These methods can be broadly categorized into two categories: threshold-based methods and typical-set-based methods. Threshold-based methods \cite{sun2021react,10.1007/978-981-99-0617-8_9} employ a pre-defined threshold to clip extremely high activations or employ a low-pass filter to exclude activations. The typical-set-based method \cite{zhu2022boosting} exhibits higher efficacy by leveraging the typical features of each channel for OOD detection. However, existing rectified-activation-based methods overlook channel characteristics, which leads to inaccurate typical set estimation and suboptimal activation rectification, ultimately degrading detection performance.
\section{Preliminaries}
\subsection{Out-of-Distribution Detection}
We present an analysis of Out-of-Distribution (OOD) detection from the perspective of hypothesis testing. Consider a $C$-class classification task with a training set $\mathcal{D}_{in}^{train}=\left\{\left(\boldsymbol{x}_{i}, y_{i}\right)\right\}_{i=1}^{n}$, where the data points are independently and identically distributed (i.i.d.) samples drawn from a joint distribution $P(\mathcal{X}, \mathcal{Y})$. Here, $\mathcal{X}$ represents the input space, $\mathcal{Y}$ denotes the label space, $\mathcal{Y} = \{1,2,\dots,C\}$ corresponds to the set of ID classes and $n$ is the number of instances in $\mathcal{D}_{in}^{train}$. We define $P_0$ as the marginal distribution of $P(\mathcal{X}, \mathcal{Y})$ with respect to the input variable $X$. Given a test input $\hat{\boldsymbol{x}}$ sampled from a test set $\mathcal{D}^{test}$, the problem of OOD detection can be formulated as a single-sample hypothesis testing task:

\begin{equation}
\mathcal{H}_0: \hat{\boldsymbol{x}} \in P_0, \quad\textbf{vs}.\quad \mathcal{H}_1: \hat{\boldsymbol{x}} \notin P_0.
\label{hypothesis_testing}
\end{equation}

According to Eq.\eqref{hypothesis_testing}, the null hypothesis $\mathcal{H}_0$ assumes that the test input $\hat{\boldsymbol{x}}$ belongs to the in-distribution (ID). The objective of OOD detection is to establish decision criteria based on $\mathcal{D}_{in}^{train}$ to determine whether to reject $\mathcal{H}_0$. This entails defining a rejection region $\mathcal{R}$ such that for any test input $\hat{\boldsymbol{x}} \in \mathcal{D}^{test}$, the null hypothesis is rejected if $\hat{\boldsymbol{x}} \in \mathcal{R}$. Typically, the rejection region $\mathcal{R}$ is determined by a test statistic and a corresponding threshold. Let $G: \mathcal{X} \mapsto \mathbb{R}^M$ be a feature extractor pre-trained on $\mathcal{D}_{in}^{train}$, where $M$ denotes the dimension of features. Let $F: \mathbb{R}^M \mapsto \mathbb{R}^C$ be a classifier trained on $\mathcal{D}_{in}^{train}$. Test-time OOD detection methods utilize $G$ and $F$ to construct a test statistic $T (\hat{\boldsymbol{x}}; F \circ G)$. Consequently, the rejection region is given by $\mathcal{R} = \{\hat{\boldsymbol{x}} : T (\hat{\boldsymbol{x}}; F \circ G ) \leq \gamma\}$, where $\gamma$ is the threshold.
\section{Method}

\subsection{OOD detection with rectified activations}
Rectified-activation-based OOD detection methods aim to enhance the separability of ID and OOD data by rectifying activations. The representative rectified-activation-based methods can be categorized into two types: (1) threshold-based methods, including ReAct \cite{sun2021react}; (2) typical-set-based method, including BATS \cite{zhu2022boosting} and LAPS \cite{he2024exploring}.  
\subsubsection{ReAct} ReAct considers extremely high activations as abnormal activations because extremely high activations cause overconfidence in predicting OOD data. To address the extremely high activations, ReAct proposes to clip them with a pre-defined activation threshold $c$:
\begin{equation}
\text{ReAct}\left(\mathrm{\boldsymbol{z}} \right)=\min(\mathrm{\boldsymbol{z}}, c)\,,
\end{equation}
where activation $\mathrm{\boldsymbol{z}}$ denotes the output of feature extractor $G$, $c$ is set based on the percentile of ID activation distribution of $\mathcal{D}_{in}^{train}$. Then the reject region can be rewritten as $\mathcal{R} = \{\hat{\boldsymbol{x}} : T (\hat{\boldsymbol{x}}; F \circ \text{ReAct} \circ G ) \leq \gamma\}$, where $\gamma$ is the threshold.

\subsubsection{BATS} BATS rectifies the features into the feature’s typical set and then uses these typical features to calculate the OOD score. BATS is then formulated as:
\begin{equation}
\text{BATS}\left(\mathrm{\boldsymbol{z}} \right)=\left\{\begin{array}{ll}\mu +\lambda \sigma , & \text { if } \quad \mathrm{\boldsymbol{z}} \ge \mu +\lambda \sigma\,;\\
\mathrm{\boldsymbol{z}}, & \text { if } \quad \mu -\lambda \sigma <\mathrm{\boldsymbol{z}} \leq \mu +\lambda \sigma\,; \\
\mu -\lambda \sigma, & \text { if } \quad \mathrm{\boldsymbol{z}} < \mu -\lambda \sigma\,,
\label{BATS_defi}
\end{array}\right.
\end{equation}
where $\lambda$ is a tuning parameter. $\mu$ and $\sigma$ denote the mean and standard deviation of the channel-level feature distribution of the training dataset. Then the reject region can be rewritten as $\mathcal{R} = \{\hat{\boldsymbol{x}} : T (\hat{\boldsymbol{x}}; F \circ \text{BATS} \circ G ) \leq \gamma\}$.

\subsubsection{LAPS} 
LAPS rectifies features by constraining them within each channel’s typical set, defined based on high-probability regions. Given a training dataset, the mean and standard deviation of the $i$-th channel’s feature distribution are $\mu_i$ and $\sigma_i$, respectively. The typical set is defined as $[\mu_i - \lambda_i \sigma_i, \mu_i + \lambda_i \sigma_i]$, where $\lambda_i$ is computed as:  
\begin{equation}\label{LAPS}
\begin{split}
    \lambda_i^1 &= \lambda + m(\bar{\mu} - \mu_i) + n(\bar{\sigma} - \sigma_i), \\
    \lambda_i^2 &= \lambda - m(\bar{\mu} - \mu_i) + n(\bar{\sigma} - \sigma_i).
\end{split}
\end{equation}
Here, $\bar{\mu}$ and $\bar{\sigma}$ represent the global mean and standard deviation of all channels. LAPS is then formulated as:  
\begin{equation}
\label{LAPS_defi}
\text{LAPS}\left(\mathrm{\boldsymbol{z}_i}\right) =
\begin{cases}
\mu_i + \lambda_i^1 \sigma_i, & \text{if } \mathrm{\boldsymbol{z}_i} \geq \mu_i + \lambda_i^1 \sigma_i, \\[1mm]
\mu_i - \lambda_i^2 \sigma_i, & \text{if } \mathrm{\boldsymbol{z}_i} \leq \mu_i - \lambda_i^2 \sigma_i, \\[1mm]
\mathrm{\boldsymbol{z}_i}, & \text{otherwise},
\end{cases}
\end{equation}
where $\mathrm{\boldsymbol{z}_i}$ denotes the feature from the $i$-th channel. The reject region is then given by $\mathcal{R} = \{\hat{\boldsymbol{x}} : T (\hat{\boldsymbol{x}}; F \circ \text{LAPS} \circ G ) \leq \gamma\}$, where $\gamma$ is the threshold.

\subsection{OOD Detection with TSRE}
Existing rectified-activation-based methods have notable limitations. Threshold-based methods, such as ReAct \cite{sun2021react}, rectify activations using globally defined thresholds, without accounting for the typicality of features. Typical-set-based methods, including BATS \cite{zhu2022boosting} and LAPS \cite{he2024exploring}, rectify activations based on estimated typical sets. Both methods neglect the distinct characteristics of channels, which leads to inaccurate estimation of the typical feature set. Furthermore, existing methods commonly assume that feature activations follow a Gaussian distribution, which is frequently violated in the true activation distribution on the training data.
These limitations call for a novel rectification strategy that takes into account channel-specific characteristics and distributional skewness.

Motivated by the above analysis, we propose a novel method named \textbf{TSRE}. Our method introduces a channel-aware typical set refinement strategy using discriminability, activity and skewness. The following subsections present the design of TSRE.

\subsubsection{Incorporating Channel Discriminability}
According to LAPS \cite{he2024exploring}, channels exhibit varying capabilities in detecting Out-of-Distribution data. Inspired by DDCS \cite{yuan2024discriminability}, such discriminative ability can be regarded as a channel-specific characteristic. Therefore, we incorporate channel discriminability into the estimation of typical sets. To estimate channel discriminability, we quantify the discriminative capacity of the $k$-th channel using two complementary properties: inter-class similarity ($S_{k}$) and the inter-class variance ($V_{k}$). Channels with low inter-class similarity and high inter-class variance are more discriminative, as they induce greater separation between classes. To estimate $S_k$ and $V_k$, we compute the prototypes of the classes.\\
\textbf{ID prototype Estimation}: We define the class prototypes ${H} = \{h^{1}, h^{2}, \dots, h^{C}\}$ based on the training set $\mathcal{D}_{\text{in}}^{\text{train}}$. For a given ID class $c$, let $x_{i}$ denote the $i$-th training sample from class $c$, and $n^{c}$ denote the total number of samples in that class. The prototype for class $c$ is then defined as:

\begin{equation}
h^{c} =\frac{1}{n^{c} } \sum_{i=1}^{n^{c}}  F(x_{i}),
\label{class_prototype}
\end{equation}
where $F$ denotes the feature extractor, and $h^c \in \mathbb{R}^M$ is the class prototype with $M$ denoting the feature dimension. Then, we calculate the average similarity $S_{k}$ between ID prototypes as follows,

\begin{equation}
S_{k} =\frac{1}{C(C-1)} \sum_{i=1}^{C} \sum_{j=1,j\ne i}^{C} \delta (h_{k}^{i} ,h_{k}^{j} )\,,
\end{equation}
where $\delta (\cdot ,\cdot )$ denotes cosine similarity, and $i,j\in \left \{1,2,\cdots,C  \right \} $ represent two different classes. Then, we calculate the inter-class variance ($V_{k}$) as follows,
\begin{equation}
V_{k} =\frac{1}{C} \sum_{i=1}^{C} (h_{k}^{i} -\bar{h_{k}} )^{2}\,,
\end{equation}
where $\bar{h_{k}} =\frac{1}{C} \sum_{i=1}^{C} h_{k}^{i} $ represents the mean of ID prototypes for the $k$-th channel. Based on the inter-class similarity $S_k$ and inter-class variance $V_k$ defined above, channel's discriminative capacity $D_{k}$ is computed as:
\begin{equation}
D_{k} = a \cdot S_{k} - (1 - a) \cdot V_{k},
\label{eq:raw_discriminative_score}
\end{equation}
where $a \in [0,1]$ balances the similarity and variance. Lower $D_{k}$ indicate stronger discriminability.

\subsubsection{Incorporating Channel Activity}

While discriminability focuses on inter-class variation based on class prototypes, we further incorporate the overall activity level of each channel to refine the typical set estimation. Specifically, channels with higher average absolute activation across all class prototypes are considered more active and informative. Channels with higher activity are more influential in OOD detection. We quantify the activity of each channel as the mean absolute activation across all class prototypes:
\begin{equation}
A^{\text{raw}}_k = \frac{1}{C} \sum_{i=1}^{C} |h_{k}^{i}|,
\end{equation}
To emphasize the most active channels, we retain only those with activity above a predefined percentile threshold. Let $\tau_p$ denote the activity value at the $p$-th percentile. The final activity is defined as:
\begin{equation}
A_k =
\begin{cases}
A^{\text{raw}}_k, & \text{if } A^{\text{raw}}_k \geq \tau_p \\
0, & \text{otherwise}.
\end{cases}
\end{equation}

The computed $A_k$ values scale each channel’s contribution when defining the typical set and computing OOD scores. This rectification amplifies the impact of salient channels while suppressing less informative activations.

Based on the computed discriminability ($D_k$) and activity ($A_k$), we adaptively determine the scaling factor $\lambda_k$ as:
\begin{equation}
\lambda_k = \lambda + \omega \cdot D_k \cdot (\bar{\mu} - \mu_k + \bar{\sigma} - \sigma_k) + A_k.
\label{lambda_update}
\end{equation}
where $\omega$ is a hyperparameter, and $(\bar{\mu} - \mu_k)$ and $(\bar{\sigma} - \sigma_k)$ account for the deviation of the $k$-th channel’s statistics from the global mean and standard deviation. This formulation allows channels with higher discriminability and activity to receive broader typical sets, promoting flexibility for activations. Given the channel-wise mean $\mu_k$ and standard deviation $\sigma_k$, the typical set for channel $k$ is then defined as:
\begin{equation}
\text{TS}_k = [\mu_k - \lambda_k \cdot \sigma_k, \, \mu_k + \lambda_k \cdot \sigma_k],
\label{}
\end{equation}
This equation specifies a symmetric interval around $\mu_k$ that defines the typical activation range for channel $k$, used to rectify anomalous activations into the typical set.

\subsubsection{Incorporating Channel Skewness}

\begin{figure}[!htbp]
    \centering
    % 在单栏中，可以适当调整图片宽度，比如 0.8 倍单栏宽
    \includegraphics[width=0.9\columnwidth]{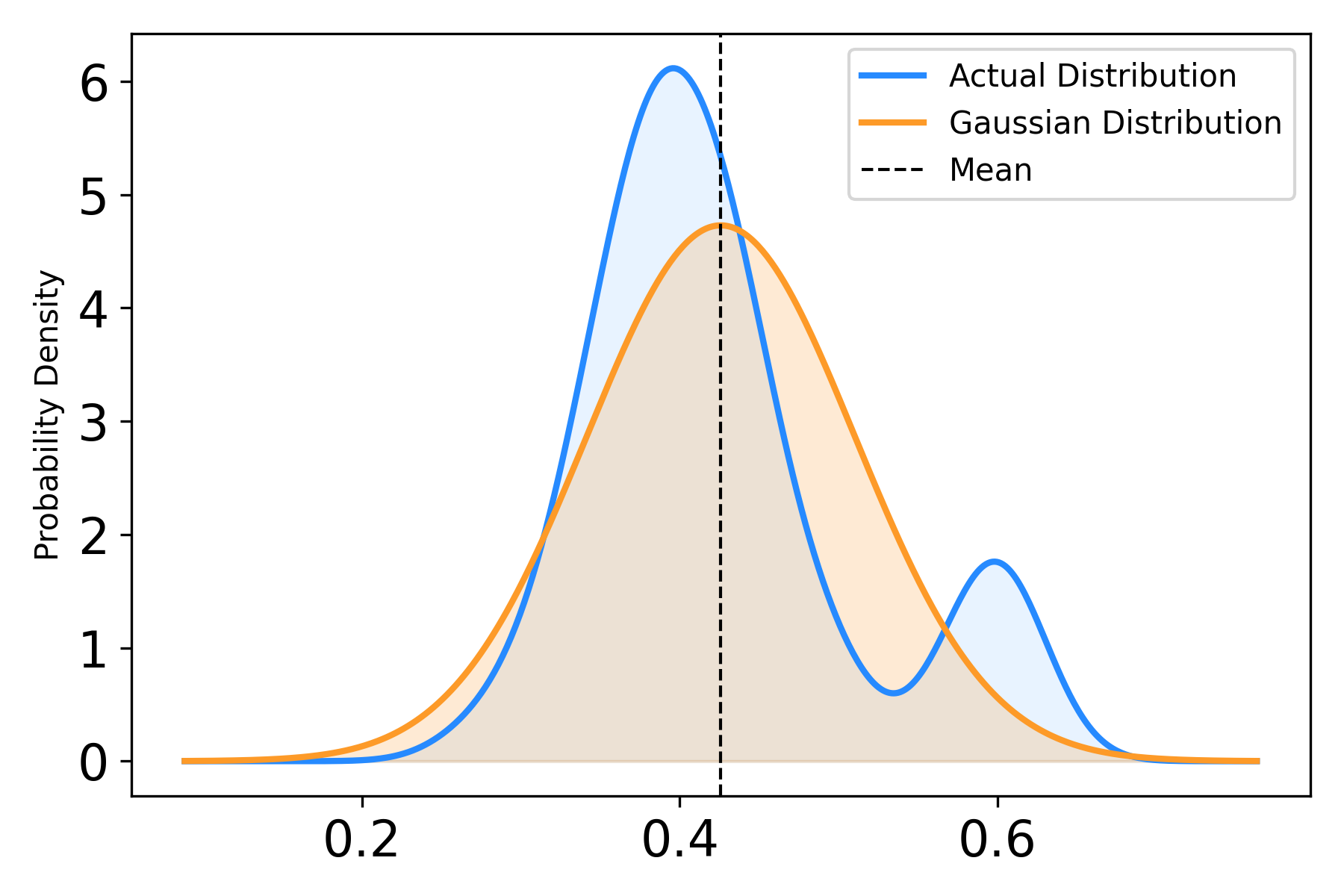}
    \caption{Significant deviation between the actual activation and the Gaussian assumption.}
    \label{fig:skewness_visualization}
\end{figure}

% 第一段描述性文字
As shown in Fig. \ref{fig:skewness_visualization}, feature activations often exhibit significant skewness across channels, violating the Gaussian assumptions commonly adopted by typical set methods such as LAPS. The skewness can distort the typical set boundaries, leading to the inclusion of outlier activations as typical or the exclusion of valid in-distribution activations, thereby degrading OOD detection performance.
% Figure1
% 使用常规的 figure 环境来插入图片
% [!htbp] 是一个灵活的位置建议符

To address the problem, we compute skewness for each channel $k$ by evaluating the third standardized moment across class prototypes:
\begin{equation}
S_k = \frac{1}{C} \sum_{c=1}^{C} \left( \frac{h_{k}^{c} - \mu_k}{\sigma_k} \right)^3,
\end{equation}
where $h_{k}^{c}$ denotes the activation of channel $k$ in the prototype of class $c$, and $\mu_k$, $\sigma_k$ are the mean and standard deviation of channel $k$ across all class prototypes. We then adjust the initial typical set boundaries by translating them according to the estimated skewness:
\begin{equation}
\text{TS}_k = \left[ \mu_k - \lambda_k \sigma_k - S_k, \, \mu_k + \lambda_k \sigma_k - S_k \right].
\end{equation}
This refinement repositions the typical set to better align with the high-probability region of the feature distribution while minimizing the influence of asymmetric responses. Channels with significant skewness thus receive appropriate boundary shifts, enhancing the robustness of OOD detection.

\subsubsection{OOD detection with rectified activation}

Based on the computed discriminability, activity, and skewness, we apply the previously defined adaptive typical set boundaries. Feature activations falling outside the typical set are rectified using the following truncation operator: \\
Let the lower and upper boundaries for the $k$-th channel be defined as:
\begin{align}
    l_k &= \mu_k - \lambda_k \cdot \sigma_k - S_k, \\
    u_k &= \mu_k + \lambda_k \cdot \sigma_k - S_k.
\end{align}
The TSRE function is then formulated as a clipping operation:
% 然后再给出 TSRE 的定义:
\begin{equation}
\label{TSRE}
\text{TSRE}(z_k) = 
\begin{cases}
    u_k, & \text{if } z_k \geq u_k, \\
    l_k, & \text{if } z_k \leq l_k, \\
    z_k, & \text{otherwise}.
\end{cases}
\end{equation}

TSRE is compatible with various downstream OOD scoring functions. By default, we adopt the energy score and further evaluate TSRE using alternative scores in the generalization analysis.

Given a test input $x^{\text{test}}$, we first extract its feature representation and rectify the channel activations using the proposed TSRE mechanism. The energy score is then computed as:
\begin{equation}
S_{\text{energy}}(x^{\text{test}}) = -\log \sum_{c=1}^C \exp\left(F \circ \text{TSRE}(G(x^{\text{test}}))\right)_c,
\end{equation}
where $G$ denotes the feature extractor and $F$ denotes the classifier, and $\circ$ represents function composition. The binary decision function is defined as:
\begin{equation}
\label{tria_energy_decision}
D_{\gamma}(x^{\text{test}}) = 
\begin{cases}
\text{ID}, & \text{if } S_{\text{energy}}(x^{\text{test}}) \geq \gamma, \\
\text{OOD}, & \text{if } S_{\text{energy}}(x^{\text{test}}) < \gamma.
\end{cases}
\end{equation}

The threshold $\gamma$ is selected to achieve a high ID acceptance rate in ID data(e.g. 95\%). By rectifying activations according to channel-aware statistical properties including discriminability, activity, and skewness, TSRE substantially improves the separation between ID and OOD data. This adaptive rectification enhances the reliability of OOD detection.

% TABLE 1
% --- 第一个表格: MobileNet-V2 on ImageNet-1K ---
\begin{table*}[tp]
    \centering
    % 使用 \resizebox 确保表格宽度合适
    \resizebox{\textwidth}{!}{
        % 内部使用常规的 tabular 环境
        \begin{tabular}{l cc cc cc cc cc}
            \toprule 
            & \multicolumn{10}{c}{\textbf{MobileNet-V2 on ImageNet-1K}} \\
            \cmidrule(lr){2-11} 
            & \multicolumn{2}{c}{iNaturalist} & \multicolumn{2}{c}{SUN} & \multicolumn{2}{c}{Places} & \multicolumn{2}{c}{Textures} & \multicolumn{2}{c}{Avg} \\
            Method & FPR95$\,\downarrow$ & AUROC$\,\uparrow$ & FPR95$\,\downarrow$ & AUROC$\,\uparrow$ & FPR95$\,\downarrow$ & AUROC$\,\uparrow$ & FPR95$\,\downarrow$ & AUROC$\,\uparrow$ & FPR95$\,\downarrow$ & AUROC$\,\uparrow$ \\
            \midrule
            MSP & 64.29 & 85.32 & 77.02 & 77.10 & 79.23 & 76.27 & 73.51 & 77.30 & 73.51 & 79.00 \\
            ODIN & 55.39 & 87.62 & 54.07 & 85.88 & 57.36 & 84.71 & 49.96 & 85.03 & 54.20 & 85.81 \\
            Energy & 59.50 & 88.91 & 62.65 & 84.50 & 69.37 & 81.19 & 58.05 & 85.03 & 62.39 & 84.91 \\
            ReAct & 45.27 & 92.40 & 53.29 & 87.58 & 61.04 & 84.39 & 41.13 & 90.85 & 50.18 & 88.80 \\
            DICE & 43.09 & 90.83 & 38.69 & 90.46 & 53.11 & 85.81 & \textbf{32.80} & 91.30 & 41.92 & 89.60 \\
            BATS & 50.63 & 91.26 & 57.36 & 86.30 & 64.46 & 83.06 & 40.00 & 91.14 & 53.11 & 87.94 \\
            LAPS & 18.82 & 96.76 & 30.07 & 92.98 & 39.70 & 90.10 & 51.37 & 88.29 & 34.99 & 92.03 \\
            TSRE & \textbf{11.11} & \textbf{98.03} & \textbf{24.88} & \textbf{94.44} & \textbf{35.16} & \textbf{91.32} & {39.27} & \textbf{92.08} & \textbf{27.60} & \textbf{93.97} \\
            \bottomrule
        \end{tabular}
    }
    % 为第一个表格创建特定的 caption 和 label
    \caption{
        OOD detection performance using MobileNet-V2 with ImageNet-1K as the ID dataset. Comparison between TSRE and other post hoc baselines. The best results are in bold. $\uparrow$ indicates higher is better, and $\downarrow$ indicates lower is better.
    }
    \label{table:results_mobilenet}
\end{table*}

% --- 第二个表格: WideResNet-40-2 on CIFAR-100 ---
\begin{table*}[tp]
    \centering
    \resizebox{\textwidth}{!}{
        \begin{tabular}{l cc cc cc cc cc}
            \toprule
            & \multicolumn{10}{c}{\textbf{WideResNet-40-2 on CIFAR-100}} \\
            \cmidrule(lr){2-11}
            & \multicolumn{2}{c}{iSUN} & \multicolumn{2}{c}{LSUN-resize} & \multicolumn{2}{c}{LSUN-crop} & \multicolumn{2}{c}{Textures} & \multicolumn{2}{c}{Avg} \\
            Method & FPR95$\,\downarrow$ & AUROC$\,\uparrow$ & FPR95$\,\downarrow$ & AUROC$\,\uparrow$ & FPR95$\,\downarrow$ & AUROC$\,\uparrow$ & FPR95$\,\downarrow$ & AUROC$\,\uparrow$ & FPR95$\,\downarrow$ & AUROC$\,\uparrow$ \\
            \midrule
            MSP   & 83.81 & 69.95 & 80.40 & 75.99 & 61.26 & 85.72 & 85.16 & 72.77 & 77.66 & 76.11 \\
            ODIN  & 89.08 & 72.43 & 77.23 & 80.83 & 21.62 & 95.81 & 79.08 & 72.32 & 66.75 & 80.35 \\
            Energy& 68.74 & 83.89 & 72.41 & 82.98 & 19.27 & 96.64 & 82.55 & 76.93 & 60.74 & 85.11 \\
            ReAct & 72.71 & 80.47 & 71.80 & 84.03 & 23.34 & 95.38 & 71.86 & 82.07 & 59.93 & 85.49 \\
            DICE & 72.48 & 81.84 & 71.99 & 82.41 & 38.60 & 93.42 & 78.03 & 76.82 & 65.28 & 83.62 \\
            BATS  & 73.12 & 79.62 & 69.41 & 85.79 & 19.11 & 96.36 & \textbf{68.28} & \textbf{82.96} & 57.48 & 86.18 \\
            LAPS  & 69.89 & 82.80 & 68.72 & 86.10 & 15.66 & 96.96 & 70.39 & 81.93 & 56.17 & 86.95 \\
            TSRE  & \textbf{58.87} & \textbf{88.87} & \textbf{65.07} & \textbf{87.03} & \textbf{13.62} & \textbf{97.51} & 70.64 & 81.12 & \textbf{52.05} & \textbf{88.63} \\
            \bottomrule
        \end{tabular}
    }
    % 为第二个表格创建特定的 caption 和 label
    \caption{
        OOD detection performance using WideResNet-40-2 with CIFAR-100 as the ID dataset. This table shows results on a different architecture and dataset pair compared to Table~\ref{table:results_mobilenet}. The best results are in bold.
    }
    \label{table:results_wideresnet}
\end{table*}
\section{Experiments}
In this section, we first introduce our experimental setup. Then, we evaluate our method on both the large-scale ImageNet-1K benchmark and the CIFAR-100 benchmarks. We also evaluate its performance across different architectures and OOD scoring functions. We conduct ablation studies to verify the effectiveness of the proposed modules. Moreover, we perform parameter sensitivity analysis to examine the impact of key hyperparameters. 

The code is available in our supplementary material.
\subsection{Set Up}
\noindent\textbf{Datasets.}   
For evaluating OOD detection on the small-scale benchmark, we use CIFAR-100 \cite{Krizhevsky_2009_17719} as the in-distribution dataset with the standard split of 50,000 training images and 10,000 test images. We consider four Out-of-Distribution datasets: LSUN-resize \cite{DBLP:journals/corr/YuZSSX15}, LSUN-crop \cite{DBLP:journals/corr/YuZSSX15}, iSUN \cite{DBLP:journals/corr/XuEZFKX15}, and Textures \cite{6909856}.  
For the large-scale benchmark, we use ImageNet-1K \cite{DBLP:journals/corr/abs-2105-01879} as the in-distribution dataset. The four Out-of-Distribution datasets include the fine-grained dataset iNaturalist \cite{DBLP:journals/corr/HornASSAPB17}, the scene recognition datasets SUN \cite{5539970} and Places \cite{7968387}, and the texture dataset Textures \cite{6909856}. All OOD datasets are selected to ensure non-overlapping categories with the in-distribution dataset.

\noindent\textbf{Backbones.}  
For the CIFAR-100 benchmark, we employ WideResNet-40-2 \cite{zagoruyko2016wide} pre-trained on CIFAR-100 as the backbone.  
For the ImageNet-1K benchmark, we use MobileNet-V2 \cite{DBLP:journals/corr/abs-1801-04381}, ResNet-18 \cite{DBLP:journals/corr/HeZRS15}, and ResNet-50 \cite{DBLP:journals/corr/HeZRS15}, all pre-trained on ImageNet-1K.

\noindent\textbf{Baselines.}  
We consider various competitive OOD detection methods as baselines, including Maximum Softmax Probability (MSP) \cite{DBLP:conf/iclr/HendrycksG17}, ODIN \cite{DBLP:conf/iclr/LiangLS18}, Energy \cite{DBLP:conf/nips/LiuWOL20}, ReAct \cite{sun2021react}, DICE \cite{sun2022dice}, BATS \cite{zhu2022boosting}, and LAPS \cite{he2024exploring}. All methods operate post-hoc using pre-trained networks. 

\noindent\textbf{Metrics.}  
FPR95: the false positive rate of OOD examples when the true positive rate of ID examples is fixed at 95\%. Lower FPR95 indicates better OOD detection performance.  
AUROC: the area under the receiver operating characteristic curve (ROC). Higher AUROC indicates better detection performance. Additional implementation and training details are provided in Appendix.

%%%\noindent\textbf{Environment details.}  
%All experiments are conducted on a hardware platform equipped with an NVIDIA RTX 4090 GPU (24 GB), running Ubuntu and CUDA 12.2. Models are implemented in PyTorch 2.6.0 and trained for 100 epochs using stochastic gradient descent (SGD) with a momentum of 0.9, a weight decay of $5 \times 10^{-4}$, and an initial learning rate of 0.001. The batch size is fixed at 256 for training and 100 for Out-of-Distribution detection.
% Table2
% 使用 table* 环境横跨双栏，[tp] 让LaTeX自由选择最佳位置
% --- 第一个表格: ResNet-50 ---
% 使用 table* 环境横跨双栏
\begin{table*}[tp]
    \centering
    % 使用 \resizebox 将表格宽度调整为页面文本宽度
    \resizebox{\textwidth}{!}{
        % 内部使用常规的 tabular 环境
        \begin{tabular}{l cc cc cc cc cc}
            \toprule 
            % --- ResNet-50 on ImageNet-1K ---
            & \multicolumn{10}{c}{\textbf{ResNet-50 on ImageNet-1K}} \\
            \cmidrule(lr){2-11} 
            & \multicolumn{2}{c}{iNaturalist} & \multicolumn{2}{c}{SUN} & \multicolumn{2}{c}{Places} & \multicolumn{2}{c}{Textures} & \multicolumn{2}{c}{Avg} \\
            Method & FPR95$\,\downarrow$ & AUROC$\,\uparrow$ & FPR95$\,\downarrow$ & AUROC$\,\uparrow$ & FPR95$\,\downarrow$ & AUROC$\,\uparrow$ & FPR95$\,\downarrow$ & AUROC$\,\uparrow$ & FPR95$\,\downarrow$ & AUROC$\,\uparrow$ \\
            \midrule
            MSP & 54.99 & 87.74 & 70.83 & 80.86 & 73.99 & 79.76 & 68.00 & 79.61 & 66.95 & 81.99 \\
            ODIN & 47.66 & 89.66 & 60.15 & 84.59 & 67.89 & 81.78 & 50.23 & 85.62 & 56.48 & 85.41 \\
            Energy & 55.72 & 89.95 & 59.26 & 85.89 & 64.92 & 82.86 & 53.72 & 85.99 & 58.41 & 86.17 \\
            ReAct & 19.99 & 96.31 & 29.60 & 93.42 & 39.70 & 90.95 & 41.42 & 91.62 & 32.68 & 93.08 \\
            DICE & 24.30 & 94.96 & 34.02 & 91.44 & 45.20 & 88.29 & \textbf{30.55} & 90.88 & 33.52 & 91.39 \\
            BATS & 24.98 & 95.51 & 25.68 & 94.27 & 37.34 & 91.11 & 32.62 & \textbf{93.47} & 30.16 & 93.59 \\
            LAPS & 12.72 & 97.50 & 15.81 & 96.18 & 24.71 & 93.64 & 41.49 & 91.81 & 23.68 & 94.78 \\
            TSRE & \textbf{11.19} & \textbf{97.79} & \textbf{14.36} & \textbf{96.34} & \textbf{23.52} & \textbf{93.80} & 34.34 & 93.18 & \textbf{20.85} & \textbf{95.28} \\
            \bottomrule
        \end{tabular}
    }
    % 更新 caption 和 label，使其具有唯一性
    \caption{
        Comparison of OOD detection performance on \textbf{ResNet-50}, with ImageNet-1K as the ID dataset. All methods are post hoc and applied on pre-trained models. The best results are in bold. $\uparrow$ indicates higher is better, and $\downarrow$ indicates lower is better.
    }
    \label{table:imagenet_resnet50}
\end{table*}

% --- 第二个表格: ResNet-18 ---
% 同样使用 table* 环境
\begin{table*}[tp]
    \centering
    \resizebox{\textwidth}{!}{
        \begin{tabular}{l cc cc cc cc cc}
            \toprule 
            % --- 第二个实验设置 ---
            & \multicolumn{10}{c}{\textbf{ResNet-18 on ImageNet-1K}} \\
            \cmidrule(lr){2-11}
            & \multicolumn{2}{c}{iNaturalist} & \multicolumn{2}{c}{SUN} & \multicolumn{2}{c}{Places} & \multicolumn{2}{c}{Textures} & \multicolumn{2}{c}{Avg} \\
            Method & FPR95$\,\downarrow$ & AUROC$\,\uparrow$ & FPR95$\,\downarrow$ & AUROC$\,\uparrow$ & FPR95$\,\downarrow$ & AUROC$\,\uparrow$ & FPR95$\,\downarrow$ & AUROC$\,\uparrow$ & FPR95$\,\downarrow$ & AUROC$\,\uparrow$ \\
            \midrule
            MSP & 58.21 & 87.06 & 73.45 & 78.90 & 75.90 & 78.06 & 71.22 & 78.43 & 69.70 & 80.61 \\
            ODIN & 46.13 & 91.66 & 59.83 & 85.03 & 66.44 & 82.64 & 50.53 & 86.65 & 55.73 & 86.50 \\
            Energy & 56.47 & 89.75 & 59.73 & 86.04 & 64.97 & 83.56 & 53.19 & 86.25 & 58.59 & 86.40 \\
            ReAct & 35.20 & 93.97 & 40.58 & 91.64 & 48.57 & 88.83 & 40.35 & 91.16 & 41.18 & 91.40 \\
            DICE & 27.03 & 94.46 & 31.22 & 92.25 & 42.38 & 88.96 & \textbf{27.15} & \textbf{92.05} & 31.94 & 91.93 \\
            BATS & 39.71 & 92.68 & 40.83 & 91.45 & 49.46 & 88.15 & 37.61 & 91.96 & 41.90 & 91.06 \\
            LAPS & 20.01 & 96.47 & 22.25 & 95.31 & 31.53 & 92.28 & 42.41 & 90.68 & 29.05 & 93.69 \\
            TSRE & \textbf{17.44} & \textbf{96.95} & \textbf{15.90} & \textbf{96.45} & \textbf{26.93} & \textbf{93.24} & 44.91 & 90.16 & \textbf{26.29} & \textbf{94.20} \\
            \bottomrule
        \end{tabular}
    }
    % 为第二个表格创建新的 caption 和 label
    \caption{
        Comparison of OOD detection performance on \textbf{ResNet-18}, with ImageNet-1K as the ID dataset. This table complements the results shown in Table~\ref{table:imagenet_resnet50}. All methods are post hoc. Best results are in bold.
    }
    \label{table:imagenet_resnet18}
\end{table*}
% Table3
% 使用常规的 table 环境，并用 [!htbp] 建议放置位置
\begin{table}[!htbp]
    \centering
    
    % --- 推荐的改动 ---
    % 1. 增加行高，让表格更舒展 (1.2 表示 120% 的默认行高)
    \renewcommand{\arraystretch}{1.2}
    % 2. 使用 tabular* 将表格宽度设置为单栏宽度
    \begin{tabular*}{\columnwidth}{@{\extracolsep{\fill}}lcc}
        \toprule
        Method & FPR95$\,\downarrow$ & AUROC$\,\uparrow$ \\
        \midrule
        Energy & 62.39 & 84.91 \\
        \midrule
        w/o all (LAPS) & 34.99 & 92.03 \\
        w/o Activity & 29.01 & 93.44 \\
        w/o Skewness & 32.35 & 92.88 \\
        w/o Discriminability & 33.76 & 92.43 \\
        \midrule
        Ours (TSRE) & \textbf{27.60} & \textbf{93.97} \\
        \bottomrule
    \end{tabular*}
    
    \caption{OOD detection performance of ablation studies on ImageNet-1K with MobileNet-V2.}
    \label{table:ablation}
\end{table}

\subsection{Main Results}
% 第一段介绍性文字
We present the comprehensive evaluation of our proposed method TSRE across two challenging benchmarks: \textbf{ImageNet-1K with MobileNet-V2} and \textbf{CIFAR-100 with WideResNet-40-2}, using four widely adopted OOD datasets for each case. Results are summarized in Table \ref{table:results_mobilenet} and Table \ref{table:results_wideresnet}.

% 第二段详细分析文字
On \textbf{ImageNet-1K}, TSRE consistently outperforms all baselines across all OOD datasets, achieving the lowest average FPR95 (\textbf{27.60\%}) and the highest average AUROC (\textbf{93.97\%}). Compared with Energy \cite{liu2020energy}, which directly uses the energy score without activation rectification, TSRE significantly improves the average FPR95 from 62.39\% to 27.60\%, and AUROC from 84.91\% to 93.97\%. This improvement highlights the effectiveness of our activation rectification strategy in enhancing feature separability.

On \textbf{CIFAR-100}, similar trends are observed. TSRE achieves the best average FPR95 of \textbf{52.05\%} and AUROC of \textbf{88.63\%}. Compared with LAPS \cite{he2024exploring}, which considers global-local statistics but ignores channel-specific characteristics, TSRE further reduces the average FPR95 from 56.17\% to 52.05\% and increases AUROC from 86.95\% to 88.63\% by introducing a typical set refinement strategy based on each channel's discriminability, activity, and skewness.

In summary, TSRE demonstrates the best performance across architectures and datasets. The consistent improvement validates the effectiveness of our method, which adaptively refines the typical set to improve OOD detection.

% Table4
% 使用 table* 环境横跨双栏，[tp] 让LaTeX自由选择位置
\begin{table*}[tp]
    \centering
    
    % 使用 tabular* 环境，让表格优雅地填充页面宽度
    % @{\extracolsep{\fill}} 是实现自动填充的关键
    \begin{tabular*}{\textwidth}{@{\extracolsep{\fill}} l cc cc cc cc}
        \toprule
        % 使用 \multicolumn 创建列分组标题
        \textbf{Method} & \multicolumn{2}{c}{\textbf{Energy}} & \multicolumn{2}{c}{\textbf{ODIN}} & \multicolumn{2}{c}{\textbf{MSP}} & \multicolumn{2}{c}{\textbf{Avg}} \\
        % 使用 \cmidrule 为每个分组画下划线
        \cmidrule(lr){2-3} \cmidrule(lr){4-5} \cmidrule(lr){6-7} \cmidrule(lr){8-9}
        & FPR95$\,\downarrow$ & AUROC$\,\uparrow$ & FPR95$\,\downarrow$ & AUROC$\,\uparrow$ & FPR95$\,\downarrow$ & AUROC$\,\uparrow$ & FPR95$\,\downarrow$ & AUROC$\,\uparrow$ \\
        \midrule
        Base Score & 62.39 & 84.91 & 54.20 & 85.81 & 73.51 & 79.00 & 63.67 & 83.24 \\
        \midrule % 使用 \midrule 分隔不同的方法，比空行更清晰
        + ReAct & 50.18 & 88.80 & 48.52 &  88.75 &  70.18 & 81.92 & 56.29 & 86.49 \\
        + BATS & 53.11 & 87.94 & 51.18 & 88.10 & 70.98 &  81.41 & 58.42 & 85.82 \\
        + LAPS & 34.99 & 92.03 & 40.84 & 90.55 & 55.18 & 86.92 & 43.67 & 89.83 \\
        + TSRE & \textbf{27.60} & \textbf{93.97} & \textbf{36.49} & \textbf{92.48} & \textbf{46.79} & \textbf{87.97} & \textbf{36.96} & \textbf{91.47} \\
        \bottomrule
    \end{tabular*}
    % 标题放在表格下方是标准做法
    \caption{TSRE improves various OOD scores. We compare the performance using the same ID dataset (ImageNet-1K) and pre-trained model (MobileNet-V2).}
    \label{table:ablation_score}
\end{table*}

\subsection{Generalization Analysis}
\noindent\textbf{Generalization to Other Architectures.} We further assess the generalizability of TSRE across different model architectures, specifically ResNet-50 and ResNet-18. As shown in Table \ref{table:imagenet_resnet50} and Table \ref{table:imagenet_resnet18}, TSRE consistently outperforms strong baselines, achieving state-of-the-art performance across both architectures. When employing ResNet-50 as the backbone, TSRE reduces the average FPR95 by 2.83\% and improves the average AUROC by 0.50\% compared to LAPS. These results highlight the strong generalization capability across architectures in OOD detection tasks.

\noindent\textbf{Generalizing to other OOD scores.} Table \ref{table:ablation_score} demonstrates the generalizability of TSRE across different OOD scores. After replacing LAPS with TSRE, the average FPR95 decreased by 6.71\%. Experimental results consistently show that TSRE improves the performance of all scoring functions. These results strongly support the effectiveness and generalizability of TSRE.

\begin{figure}[htbp!]
    \centering
    % 第一行子图
    \begin{subfigure}[b]{0.49\columnwidth} % 宽度设置为略小于单栏宽度的一半
        \includegraphics[width=\textwidth]{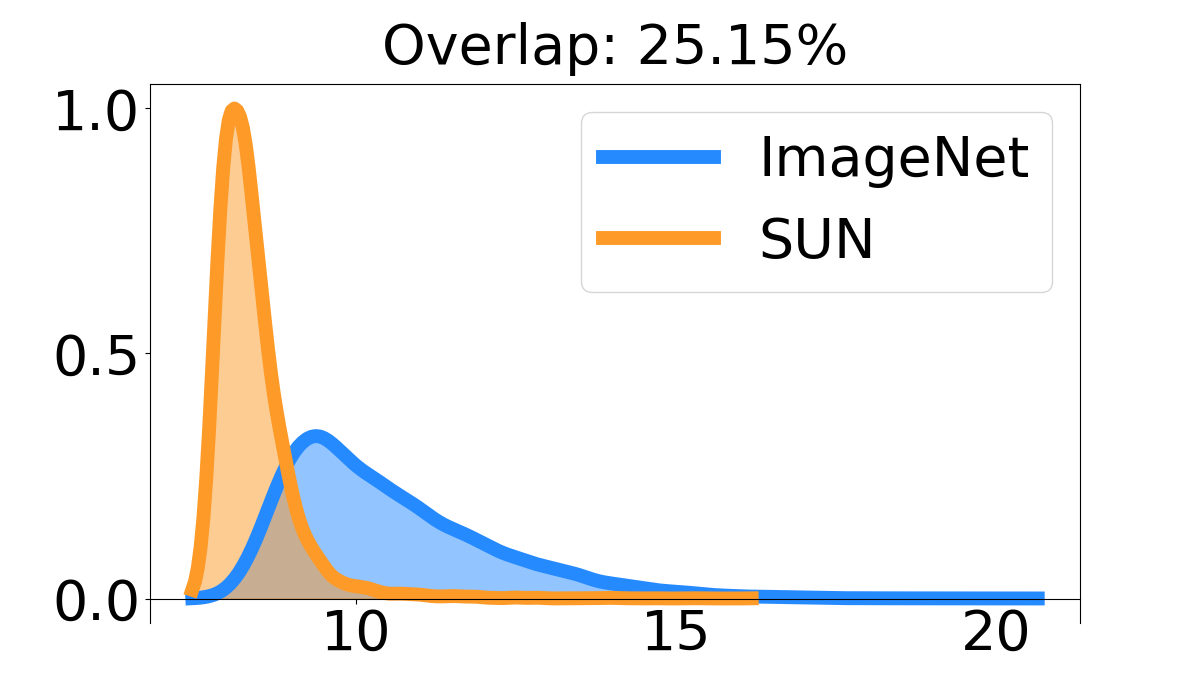}
        \caption{TSRE}
        \label{fig:ablation_a} % 为每个子图设置独立的标签
    \end{subfigure}
    \hfill % 在两张子图之间添加弹性空白，让它们分别靠向左右两边
    \begin{subfigure}[b]{0.49\columnwidth}
        \includegraphics[width=\textwidth]{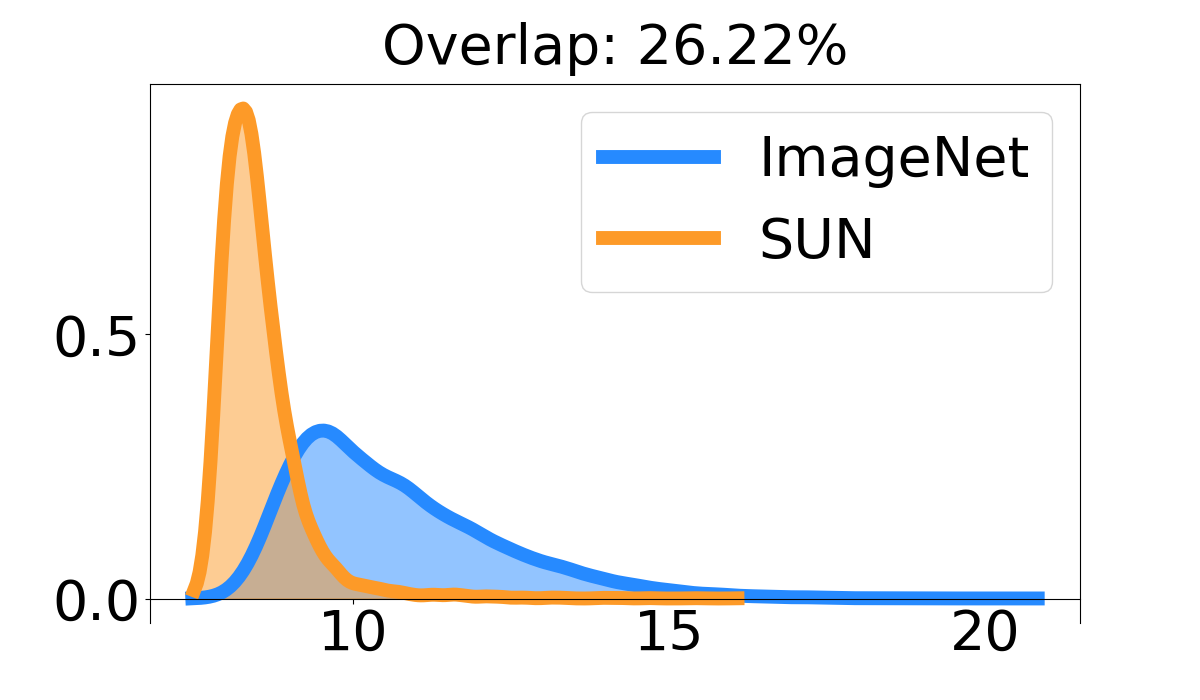}
        \caption{w/o Activity}
        \label{fig:ablation_b}
    \end{subfigure}

    % 在这里插入一个空行，或者用 \vspace 添加垂直间距，实现换行
    \vspace{3mm} 

    % 第二行子图
    \begin{subfigure}[b]{0.49\columnwidth}
        \includegraphics[width=\textwidth]{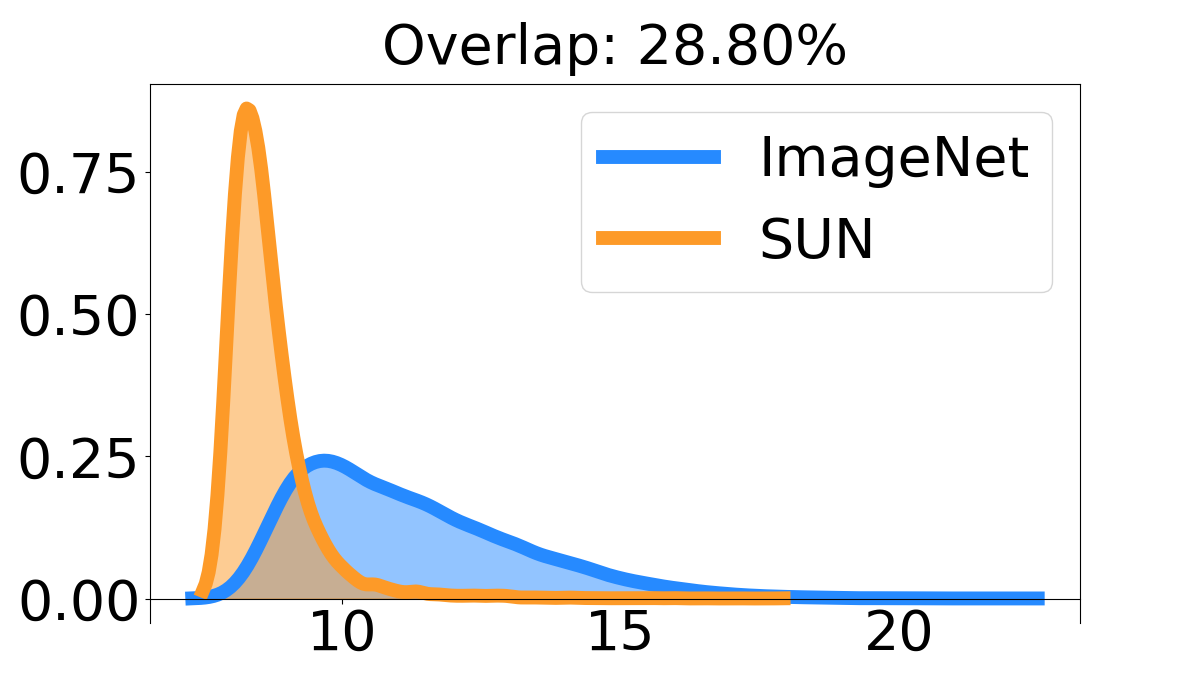}
        \caption{w/o Skewness}
        \label{fig:ablation_c}
    \end{subfigure}
    \hfill
    \begin{subfigure}[b]{0.49\columnwidth}
        \includegraphics[width=\textwidth]{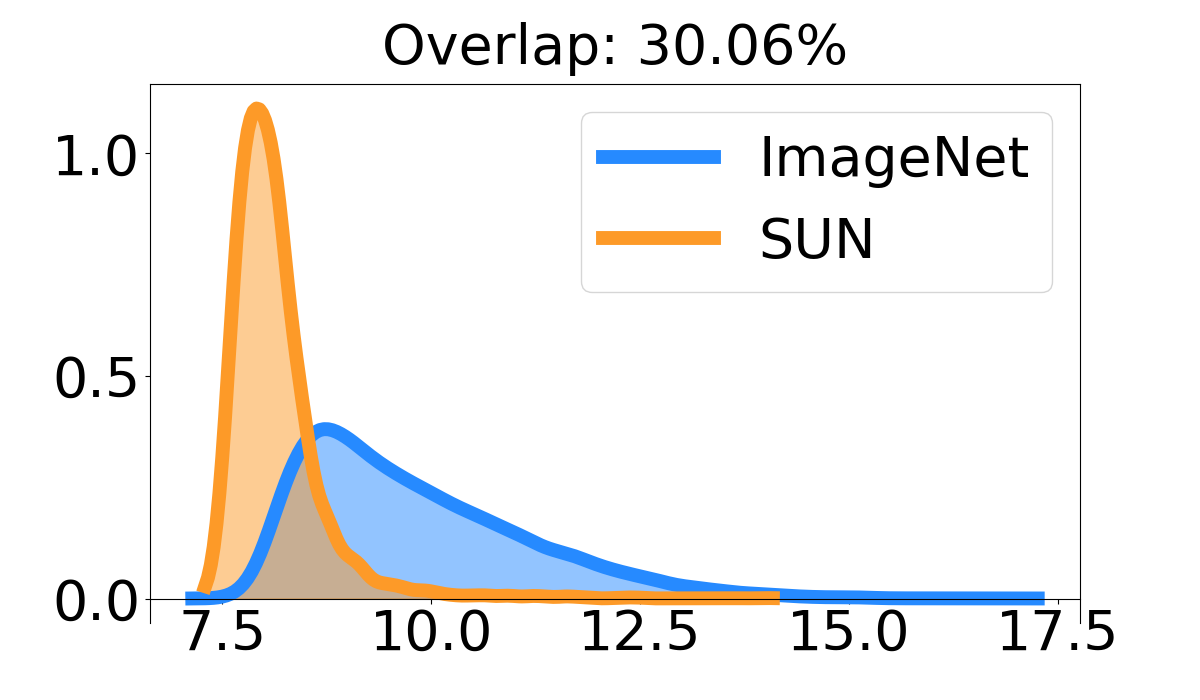}
        \caption{w/o Discriminability}
        \label{fig:ablation_d}
    \end{subfigure}
    
    % 整个 figure 的总标题和总标签
    \caption{Ablation study on TSRE. The distribution overlap between ImageNet-1K (ID) and SUN (OOD) increases when each channel-aware factor is removed, validating the contributions of discriminability, activity, and skewness to enhancing ID-OOD separability.}
    \label{fig:Ablation}
\end{figure}

\begin{figure}[h]
    \centering
    % 这个设置可以减少图片和其下方子标题之间的距离，保留它是个好习惯
    \captionsetup[subfigure]{skip=1pt}

    % 第一个子图
    \begin{subfigure}{0.75\columnwidth}
        \centering
        % 图片宽度设为100%子图宽度，即 \linewidth
        \includegraphics[width=\linewidth]{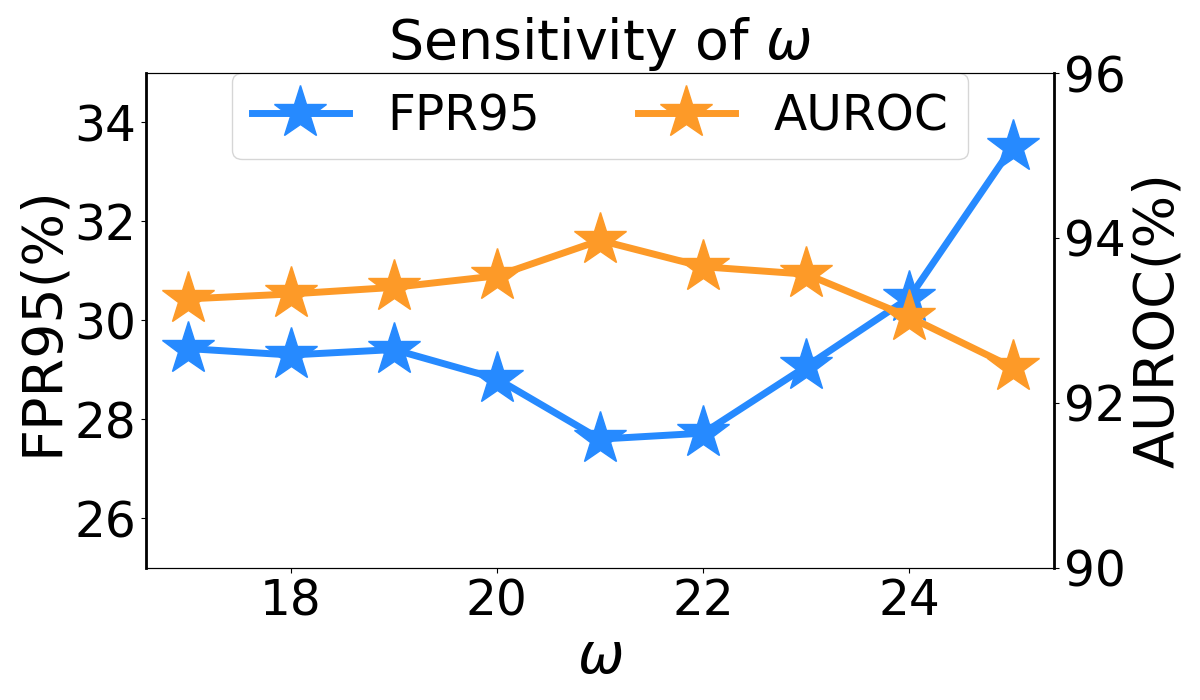}
        \caption{Sensitivity of $\omega$}
        \label{fig:sensitivity_omega}
    \end{subfigure}
    
    % \vspace{...} % 如果需要，可以在子图之间添加额外的垂直间距
    
    % 第二个子图
    \begin{subfigure}{0.75\columnwidth}
        \centering
        \includegraphics[width=\linewidth]{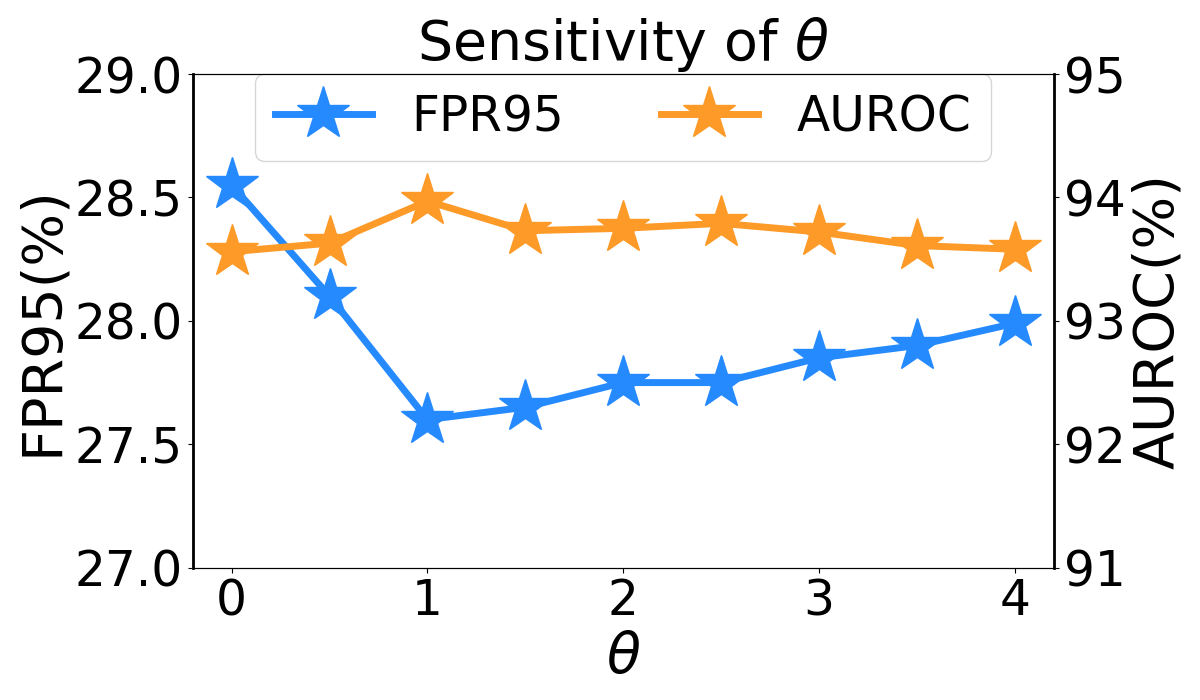}
        \caption{Sensitivity of $\theta$}
        \label{fig:sensitivity_theta}
    \end{subfigure}
    
    % 第三个子图
    \begin{subfigure}{0.75\columnwidth}
        \centering
        \includegraphics[width=\linewidth]{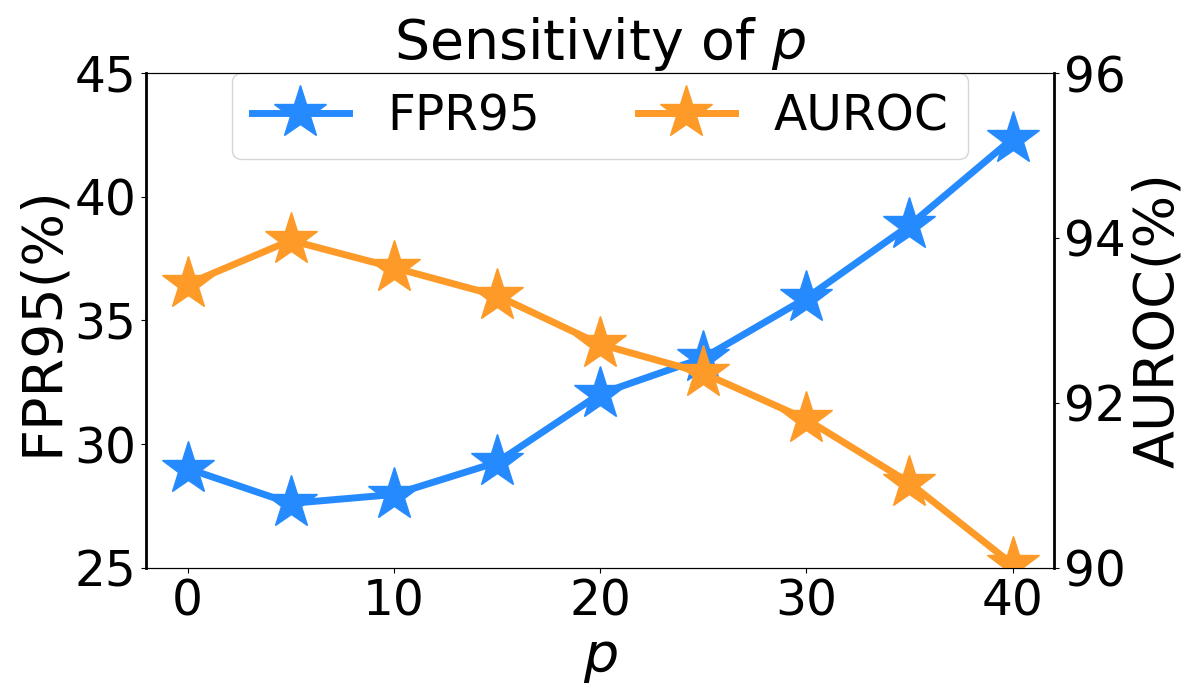}
        \caption{Sensitivity of $p$}
        \label{fig:sensitivity_p}
    \end{subfigure}
    
    % 主标题
    \caption{Sensitivity analysis of hyperparameters $\omega$, $\theta$, and $p$ on FPR95 and AUROC.}
    \label{fig:sensitivity_all}
\end{figure}

\subsection{Ablation Study}

% Figure2
% 使用常规的 figure 环境来插入图片
% [!htbp] 是一个灵活的位置建议符，强烈建议LaTeX在此处、页面顶部或底部放置
% Figure3
% 使用常规的 figure 环境，并用 [tp] 或 [!t] 让它浮动到单栏的顶部
% 使用 figure* 环境来让图片跨双栏

\begin{figure}[htbp!]
    \centering
    % 在单栏中，宽度基准应为 \columnwidth
    % 0.9\columnwidth 会让图片占据单栏80%的宽度，留出漂亮的边距
    \includegraphics[width=1.0\columnwidth]{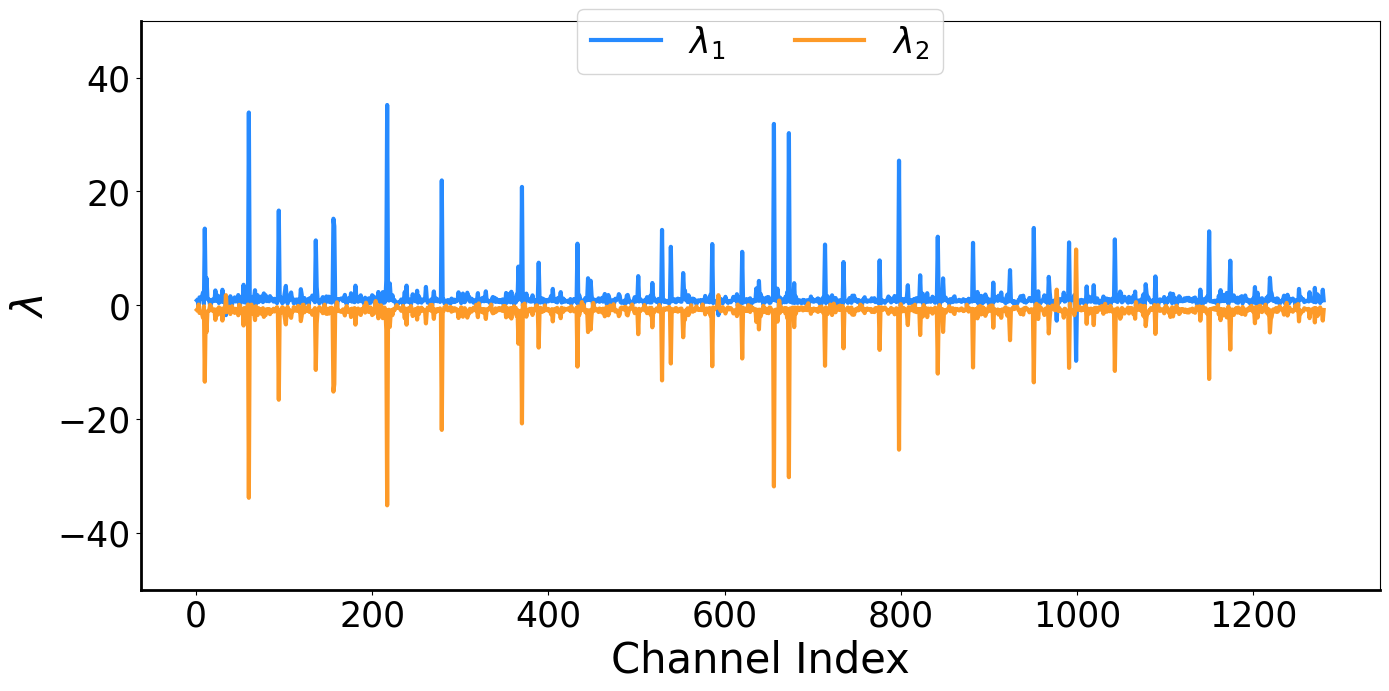}
    \caption{The distribution of scaling factors $\lambda_i$ after adaptive refinement.}
    \label{fig:lambda_distribution}
\end{figure}

Table~\ref{table:ablation} presents the results of the ablation study using Energy as the base OOD scoring function. \textbf{w/o Activity}, \textbf{w/o Skewness}, and \textbf{w/o Discriminability} correspond to variants of our method with each channel-aware factor individually removed, while \textbf{w/o all} denotes the version without any refinement, equivalent to LAPS. Fig. \ref{fig:lambda_distribution} visualizes the scaling factors $\lambda_i$ used in our initial typical set estimation based on discriminability and activity, revealing that each channel exhibits distinct characteristics. As illustrated in Fig. \ref{fig:Ablation}, removing any single component leads to increased overlap between ID and OOD feature distributions and a decline in OOD detection performance. These results demonstrate that discriminability, activity, and skewness each contribute uniquely and jointly to the effectiveness of the proposed method. Discriminability reflects the channel’s capacity to separate categories, activity emphasizes channels with high activations, and the refinement using skewness mitigates the estimation bias of the typical set caused by incorrect distributional assumptions. Together, these components enhance the separability of ID and OOD data by refining the estimation of channel-aware typical sets.

\subsection{Sensitivity Analysis}
Our method introduces three essential hyperparameters: $\omega$, $\theta$, and $p$. Fig. \ref{fig:sensitivity_all} presents the empirical sensitivity of FPR95 and AUROC with respect to these parameters. The results indicate that our settings ($\omega=21$, $\theta=1$, $p=5\%$) offer the best performance. Notably, either increasing $\omega$  or decreasing it leads to worse metrics, reflecting overly aggressive or insufficient typical set refinement. Similarly, varying $\theta$ and $p$ away from their optimal values results in performance degradation due to insufficient or excessive rectification. These results confirm that careful selection of $\omega$, $\theta$, and $p$ is essential for effective OOD detection, as it helps minimize false positive rates and maximize the area under the receiver operating characteristic curve.

\section{Conclusion}
In this work, we identify a key limitation of existing typical set refinement methods: their inability to exploit intrinsic channel-level properties, which ultimately limits their effectiveness in out-of-distribution detection. To address this challenge, we propose TSRE, a channel-aware typical set refinement framework that integrates discriminability, activity, and skewness to rectify feature activations. Our visualizations demonstrate that TSRE effectively improves activation rectification, which in turn enhances out-of-distribution detection performance. Extensive experiments demonstrated that TSRE outperforms existing methods on CIFAR and ImageNet-1K. Our experiments also demonstrate that our method generalizes effectively across different backbones and scoring functions. In the future, we will integrate TSRE into other dynamic open-world tasks.

{
    \small
    \bibliographystyle{ieeenat_fullname}
    \bibliography{main}
}


\end{document}